\documentclass[conference]{IEEEtran}
\usepackage{cite}

\usepackage{amsmath}
\usepackage{amssymb}
\usepackage{amsfonts}
\usepackage{amsthm}
\usepackage{mathtools}

\usepackage{graphicx}
\usepackage{booktabs}
\usepackage{array}
\usepackage{makecell}
\usepackage[caption=false,font=footnotesize]{subfig}
\usepackage[a4paper, left=0.65in, right=0.65in, top=0.73in, bottom=1in]{geometry}

\usepackage{algorithm}
\usepackage{algorithmic}

\usepackage{microtype}
\usepackage{textcomp}
\usepackage{xcolor}

\usepackage[
  hidelinks,
  bookmarks=false,
  draft
]{hyperref}
\theoremstyle{plain}
\newtheorem{theorem}{Theorem}[section]

\newtheorem{lemma}[theorem]{Lemma}

\theoremstyle{definition}
\newtheorem{definition}[theorem]{Definition}

\theoremstyle{remark}
\newtheorem{remark}[theorem]{Remark}

\def\BibTeX{{\rm B\kern-.05em{\sc i\kern-.025em b}\kern-.08em
    T\kern-.1667em\lower.7ex\hbox{E}\kern-.125emX}}
\begin{document}

\title{MultiTok: Variable-Length Tokenization for Efficient LLMs Adapted from LZW Compression}

\author{
\IEEEauthorblockN{Noel Elias$^{1}$, Homa Esfahanizadeh$^{{2}}$, Kaan Kale$^{3}$, Sriram Vishwanath$^{4}$, and Muriel M\'edard$^{5}$}
\IEEEauthorblockA{
$^{1}$University of Texas at Austin, Austin, TX, USA, Email: nelias@utexas.edu
}
\IEEEauthorblockA{
$^{2}$Nokia Bell Labs, Murray Hill, NJ, USA, Email: homa.esfahanizadeh@nokia-bell-labs.com
}
\IEEEauthorblockA{
$^{3}$Boğaziçi University, Istanbul, Türkiye, Email: huseyin.kale@std.bogazici.edu.tr
}
\IEEEauthorblockA{
$^{4}$Georgia Institute of Technology, Atlanta, Georgia, USA, Email: sriram@ece.gatech.edu
}
\IEEEauthorblockA{
$^{5}$Massachusetts Institute of Technology (MIT), Cambridge, USA, Email: medard@mit.edu
}
}

\maketitle

\begin{abstract}
Large language models have drastically changed the prospects of AI by introducing technologies for more complex natural language processing. However, current methodologies to train such LLMs require extensive resources including but not limited to large amounts of data, expensive machinery, and lengthy training. To solve this problem, this paper proposes a new tokenization method inspired by  universal Lempel-Ziv-Welch data compression that compresses repetitive phrases into multi-word tokens. With MultiTok as a new tokenizing tool, we show that language models are able to be trained notably more efficiently while offering a similar accuracy on more succinct and compressed training data. In fact, our results demonstrate that MultiTok achieves a comparable performance to the BERT and GPT standards as both a stand-alone tokenizer and an add-on to existing tokenizers while also providing close to 2.5x faster training with more than 30\% less training data.
\end{abstract}


\section{Introduction}
In recent times, the landscape of AI-based applications has dramatically changed with the introduction of Large language models (LLMs). With the emergence of technologies like ChatGPT, LLaMA, etc., users are able to utilize LLMs for natural language processing tasks including text generation, code generation, summarization, and even complex reasoning \cite{wan2023efficient}. However, to achieve these capabilities, LLMs must utilize massive amounts of diverse training data. In particular, most LLMs need over 1 billion tunable parameters to learn from such large amounts of data. Consequently, it often takes a notably long time to train advanced LLMs, utilizing expensive GPU machinery. In fact, according to \cite{wan2023efficient}, the time needed to train language models increases exponentially with the training data size. On the other hand, the training data size for language models has grown 3x per year since 2010 \cite{epoch2023aitrends}. Thus, these extensive resource requirements and lengthy training time for LLMs call attention to the need for more efficient techniques, motivating training on compressed data, which is our focus. 

To mitigate challenges regarding the massive training data, data compression techniques have been deployed to reduce the size of the data and minimize additional overhead. Current techniques for training data compression include methods like PCA\cite{mackiewicz1993principal}, t-SNE\cite{van2008visualizing}, and mutual information-based methods\cite{Kale2024TexShapeIT} for feature dimension reduction, among others. Specifically, such techniques work by reducing the resulting token embedding size to lower-dimensional spaces. 

Additionally, methods such as transferring knowledge across language models\cite{petroni2019language} and creating sparse representation for data \cite{dhillon2001concept} have been utilized to reduce the training cost. These methods utilize post-processing techniques to reduce model parameters. However, many of these solutions are not dynamic and still require some baseline tokenization or rely on specific linear data distributions. There also exists methods to reduce inference cost via removing unnecessary neural parameters through quantization\cite{han2015deep}. Still, these methodologies are post processing methods, require a large amount of resources for training, and are computationally costly to run. 

Tokenization is an essential component of many current large language pipelines \cite{rajaraman2024toward, mielke2021between, grefenstette1999tokenization}, which directly affects the volume of training data and training complexity. Some examples of tokenization algorithms in the literature are variants of the WordPiece \cite{wu2016google} and Byte Pair Encoding \cite{gage1994new,sennrich2015neural}. These algorithms are building blocks for generating the state-of-the-art tokens utilized in the Bidirectional Encoder Representations from Transformers (BERT) \cite{devlin2018bert} and Generative Pre-trained Transformer 2 (GPT) \cite{radford2019language} models. %However, these schemes do not solve the existing problem of large amounts of training data. 

A promising direction for addressing the challenge of large training data size is using compression algorithms on training data \cite{deletang2023language, sennrich2015neural}. Specifically, compression algorithms like Huffman coding\cite{huffman1952method} and other lossless compression algorithms\cite{deletang2023language} have been evaluated as tokenizers\cite{gajjala2020huffman, wolleb2023assessing, shu2017compressing}. In addition, implementing linear algebraic error-correcting codes have also been in-play \cite{anjum2024lipcot, basit2007efficient}. However, these schemes fall short in their performance or their computational cost with respect to traditional tokenizers. 

One of the most well-known lossless compression algorithms is the Lempel-Ziv-Welch (LZW) algorithm \cite{welch}. The LZW algorithm is a universal compression algorithm that works by compressing repeated patterns of bits within a dataset. As such, it achieves a high compression ratio and performs better with larger amounts of data \cite{gupta2017modern}. In this paper, we introduce \texttt{MultiTok}, a new tokenization method inspired by the LZW compression algorithm for the efficient LLM training tasks. We focus on LLMs where the inputted data is first tokenized utilizing an encoder, extracted into embedding vectors representing the semantic and contextual information, and finally fed into a transformer model for text classification. \texttt{MultiTok} provides a novel variable-length tokenizer in the sense that each token can represent a variable number of words. The advantages of \texttt{MultiTok} are it (i) dynamically compresses the necessary training data by $33 \%$ (ii) allows the LLM to be trained close to 2.5x times faster and (iii) maintains performance comparable to the baseline BERT and GPT token standards. Our tokenizing schemes can mark the beginning of the use of information-theoretic approaches to provide efficient, secure, and robust LLM systems. The code and data can be accessed at: \texttt{https://github.com/noelkelias/multitok}.

\section{Problem Setting}
We denote the set of all samples of a distribution by $\mathcal{X}$. Each sample $x\in\mathcal{X}$ in the training data is described as a sequence of tokens, i.e.,
\begin{equation}
    x=[x_1,\dots,x_{n(x)}],
\end{equation}
where $n(x)$ is the number of tokens in the sample $x$. For instance, each $x_i$ can be a single-word token, or alternatively can be obtained from more advanced tokenization schemes in the literature \cite{devlin2018bert}. All tokens are recorded in a dictionary $D$, where each token is then paired with a unique embedding vector for the LLM tasks. 

The focus of our paper is on building an efficient multi-word tokenization scheme $T$. Our method is based on converting the original tokens of each sample into a smaller number of tokens, by taking a consecutive number of original tokens and grouping them together into a new token, 
\begin{equation}
x=[x_1,\dots,x_{n(x)}]\rightarrow T(x)=[y_1,\dots,y_{m(x)}].
\end{equation}
Here, $m(x)<n(x)$, and the new shorter sequence of tokens is a consecutive disjoint split of tokens in $x$. 

We consider several performance goals: The utility goal is to train a competitive downstream classifier with a high generalization accuracy using the proposed tokenization scheme.  The utility is often quantified via the generalization accuracy over some validation data. The efficiency goal is two-fold: compression (having fewer tokens per sample) and convergence (reducing the training time).

\begin{definition} \label{def:compress-ratio}
The ratio of the number of tokens in the dataset, before and after applying $T$ is defined as the compression ratio,
$$r = \frac{\sum_{x\in\mathcal{X}}|T(x)|}{\sum_{x\in\mathcal{X}}|x|} =\frac{\sum_{x\in\mathcal{X}}m(x)}{\sum_{x\in\mathcal{X}} n(x)}.$$
\end{definition} 

\begin{definition} \label{def:training-time}
Let $l_i$ be the training loss that the model exhibits on the $i$-th epoch. The training time is defined as
$$
C(\epsilon)=\min\{i|l_{i+1},\dots,l_{i+10}<\epsilon\}.
$$
\end{definition}

Utility and efficiency goals are often competing requirements. In this paper, we design a tokenizer that optimizes for both goals and provides parameters that offer a utility-efficiency trade-off.  

\begin{figure*}
    \centering
    \includegraphics[width=0.9\linewidth]{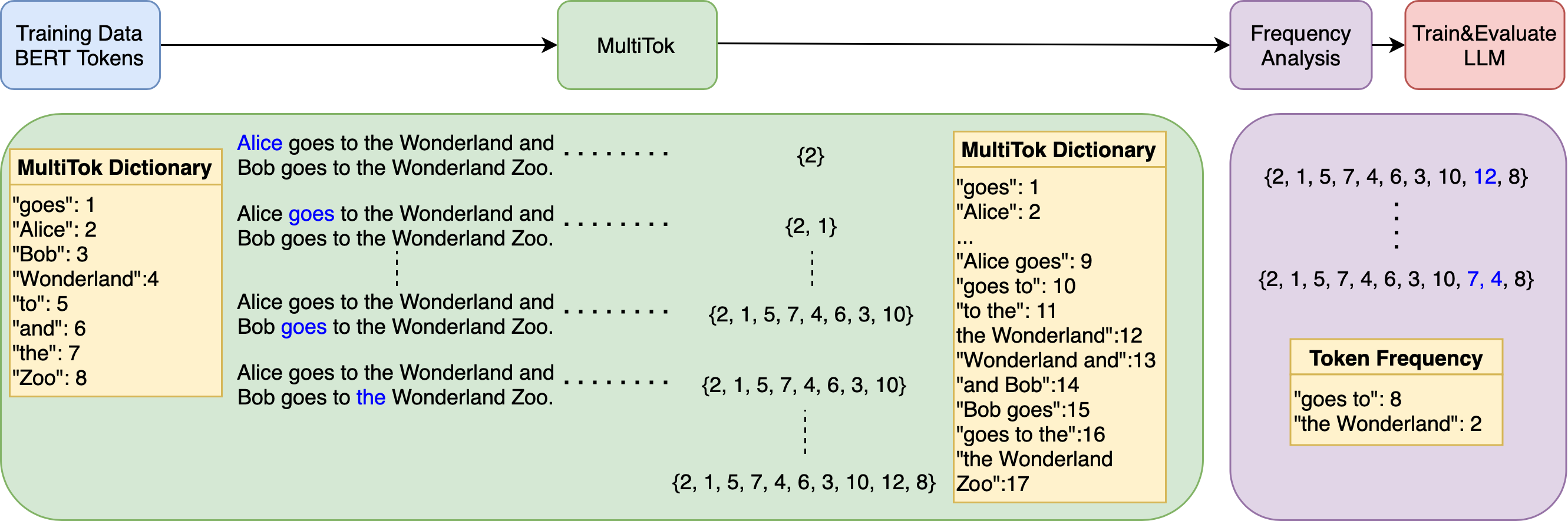}
    \caption{A toy example showing how \texttt{MultiTok} dictionary is constructed and used for tokenization.}
    \label{fig:diagram}
\end{figure*}

\section{MultiTok: A Variable-Length Tokenization}
\texttt{MultiTok} works by building a dictionary while encoding the samples into a sequence of tokens, inspired by the LZW compression scheme \cite{welch}. We first briefly explain the method: Within a training window of size  $w$, \texttt{MultiTok} looks ahead from the current word, starting from the first word of the first training sample, to identify the largest multi-word token it has previously encountered (already in the dictionary). A new token, comprising this largest known token and the next word, is then added to the dictionary. The process is repeated with the newly added word as the focus. Once all words in a given sample have been processed, the algorithm moves to the next sample, continuing until the entire dataset has been used to build a more comprehensive dictionary. This enables \texttt{MultiTok} to add more common phrases into the dictionary, making it larger, but reducing the average number of tokens needed to represent a sample in the training data. 

An overview of the \texttt{MultiTok} tokenization scheme is illustrated as Algorithm \ref{alg:multitok}.  The input of the algorithm is the training dataset $\mathcal{X}$, an initial dictionary of all possible single-word tokens $D$, and the maximum number of words that a \texttt{MultiTok} token can represent as $w$. In Line 3, we initialize the \texttt{MultiTok} vector $T(x)$ for sample $x$ and the current length of our dictionary as $idx$. Through Lines 4 and 5, we iterate through each word within $x$. For each word, we look $w$-word ahead to form a multi-word phrase of maximum length $w$, starting from the shortest phrase. When we find the smallest multi-word phrase that is not already in our dictionary $\mathcal{D}$, we will add it to the dictionary (lines 6 and 7). Then, we add the current largest known multi-word token to our \texttt{MultiTok} vector $T(x)$ and move on to encode the next available word (Line 7). If we have reached the end of our sample, we add the index of the remaining phrase to our \texttt{MultiTok} vector in Line 10. Finally, we add the vector of tokens representing the sample $x$ to the set $\mathcal{T}$.

We note that $x[i{:}j]$ is the sub-vector of $x$ with the elements from indices $i$ to $j$. Further, the operator ${+=}$ denotes appending an element to the end of a vector. During inference, for tokenization, we start from the first word of a sample, and identify the largest match in the dictionary with the length at most $w_{\text{test}}$ (called testing window). We note that $w_{\text{test}}$ can be different than $w$ (called training window) in Algorithm~\ref{alg:multitok}.

\begin{algorithm}[]
\small
 \caption{\texttt{MultiTok} Tokenization Pseudocode }\label{alg:multitok}
 \begin{algorithmic}[1]
 \renewcommand{\algorithmicrequire}{\textbf{Input:}}
 \renewcommand{\algorithmicensure}{\textbf{Output:}}
 \REQUIRE $\mathcal{X}, D, w$ 
 \ENSURE  $\mathcal{T}=\{T(x)\}_{x\in\mathcal{X}}$ 
  \STATE $\mathcal{T}=\{\}$
  \STATE $idx = |\mathcal{D}|+1$
  \FOR {$x \in \mathcal{X}$} 
  \STATE $i=1, T(x)=[{}]$
  \WHILE{$i \le |x|$}
  \FOR {$j \in [i, \text{min}(i+w,|x|)]$} 
  \IF{$x[i:j] \notin D$}
    \STATE $D(x[i:j]) = idx,\; idx=idx+1$
    \STATE $T(x) {+\!=} D(x[i:j-1]),\; i:= j$
    \STATE \textbf{break}
  \ENDIF 
    \IF{$j=|x|$}
        \STATE $T(x) \mathrel{+}= D(x[i:j])$
    \ENDIF
  \ENDFOR
  \STATE $\mathcal{T} = \mathcal{T} \cup T(x)$
  \ENDWHILE
  \ENDFOR
 \end{algorithmic}
 \end{algorithm}

We also propose a post-processing technique, where we conduct frequency analysis on the \texttt{MultiTok} dictionary. Particularly, multi-word tokens that only appear a few number of times within the encoded training sample are pruned. As a result, during the encoding, those larger tokens are replaced by their smaller constituent tokens that are more frequently used. The motivation for this step is to provide a smaller dictionary of more frequently utilized tokens for a training task rather than including multi-word tokens that might have been rarely used during tokenization. According to our experiment (see Table \ref{tab:results}), by pruning the tokens that are utilized less than twice both the accuracy and the compression ratio slightly increase by up to 6\%.

Lemma~\ref{lemma:ratio} describes how much the compression ratio increases after the proposed frequency-based post-processing.
\begin{lemma}\label{lemma:ratio}
   Let $r$ and $r'$ denote the compression ratio before and after the frequency-based post-processing. For a threshold $K$, where the tokens that appear equal or less than $K$ times are replaced with their smaller, more-frequent constituents, we have
   \begin{equation*}
       \frac{r'}{r} \leq \frac{N-\sum_{k=1}^K\left(\lambda_k-\lambda_k w\right)}{N}.
   \end{equation*}
Here, $N$ is the number of tokens in the original \texttt{MultiTok} encoded data, $\lambda_k$ is the number of tokens that appear $k$ times in the training data, and $w$ is the training window size. 
\end{lemma}

\begin{proof}
The number of tokens before the post-processing is $N$. During the post-processing, for every $k \in \{1,\dots,K\}$, the corresponding $\lambda_k$ multi-word \texttt{MultiTok} tokens that only appear $k$ times in the training data are removed. Each of these tokens are replaced 
with their constituents that are either visited more than $K$ times or are single-word tokens. The number of these constituents are upper bounded by $w$ by construction. Thus, the number of tokens after the post processing is at most $N-\sum_{k=1}^K\left(\lambda_k-\lambda_k w\right)$, and the upper bound for the ratio  of $r$ and $r'$ is concluded.
\end{proof}

\begin{remark}
We hypothesize that this post-processing step increases the accuracy of the downstream model, confirmed by our experimental results. In particular, when a token rarely occurs, the relation of that token with other tokens is less explored during training. Thus, the semantics can be less captured, degrading the performance. %The results capture this observation by demonstrating an increase in the accuracy from $64.8\%$ to $70.9\%$ when applying this post-processing technique to the \texttt{MultiTok}, 100\%, Max-Max experiment. 
\end{remark}

Finally, we present the end-to-end tokenization pipeline utilizing Algorithm~\ref{alg:multitok}  through an example in Fig.~\ref{fig:diagram}. We begin by tokenizing the phrase ``Alice goes to the Wonderland and Bob goes to the Wonderland Zoo". We first find the largest existing phrase in the inputted dictionary (left) and thus tokenize ``Alice" as 2. Then, we add ``Alice goes" to the \texttt{MultiTok} dictionary (right) as a new token with index 9. We continue this process until the second mention of the phrase ``goes to". This phrase is already in the \texttt{MultiTok} dictionary and is thus encoded as 10 instead of tokens 1 and 5. Now, we add ``goes to the" to the \texttt{MultiTok} dictionary with index 16. Afterwards, the tokenized samples undergo frequency analysis (purple). Particularly, we prune tokens like token 12 which were only used twice within the entire training data set $\mathcal{X}$. Thus, its instances in the dataset are replaced with the two tokens that it compresses, i.e., 7 and 4. The resulting tokenized sequence is inputted into the LLM for training. 

\section{Simulations Results} 
In this section, we demonstrate the empirical performance of our tokenizer, \texttt{MultiTok},  with respect to the utility and efficiency goals described by Definitions \ref{def:compress-ratio} and \ref{def:training-time}. Further, we show various goals such as utility, accuracy, and training time can compete, and we demonstrate some parameters that can offer a good trade-off among them. 

\subsection{Experimental Setup}
The general pipeline includes tokenizing the training samples with the \texttt{MultiTok}, BERT, and GPT tokenization schemes. These tokens were then assigned corresponding random vector embedding and inputted to train a selected model for the textual binary classification task.

\textbf{Training.} 
The datasets utilized in this section are the IMDB \cite{maas-EtAl:2011:ACL-HLT2011}, the sst2 \cite{socher-etal-2013-recursive}, and the AG-News \cite{Zhang2015CharacterlevelCN} datasets. The first two are collections of movie reviews, each annotated with a binary sentiment. The third is a set of news pieces from different categories. Our experiments were performed on Google Colab instances (Xeon Processors @2.3Ghz with 12GB of RAM). The selected model utilizes a random vector embedding of dimension $100$ per token. This is fed to a model including a single transformer layer \cite{vaswani2017attention}, with two attention heads, a position-wise feed-forward neural network containing two linear layers of dimensions $100$ and $200$, and a few LayerNorm \cite{ba2016layer} normalization layers of dimension $100$. Then, the nodes are flattened to an output layer for the binary classification. All models were trained for $20$ epochs utilizing batch sizes of $1000$, the Adam's optimizer, binary cross-entropy loss, and a learning rate of $0.01$. The complete model configuration and hyperparameters can be found in our code repository \texttt{https://github.com/noelkelias/multitok}.

Our goal with the simulation was to ensure we \textit{most directly} evaluate the effect of different tokenizations on a model's performance. The best way to do this is to eliminate possible confounding variables outside tokenization that could affect a model's performance. Utilizing more complex architectures in smaller scale experiments could overcompensate a model's performance for bad tokenization \cite{zhou2024larger}. In addition, the performance of more complex classification tasks are often reflective of the underlying architecture rather than the inputted tokens \cite{singh2024tokenization, rajaraman2024toward}. The alternative was to train a LLM with vast amounts of training data and billions of parameters, which would negate immediate architectural and other confounding advantages, thereby focusing more directly on inputted training data (tokenizers). For this reason, we used a simple transformer model  and tested on a basic classification dataset.

\textbf{Evaluation.} 
All our results are the averages over 20 trials with the same parameters. Specifically, we record the model's average training loss, training accuracy, and testing accuracy per epoch. In addition, the compression ratio as well as the average area under the curve (AUC) given by the receiver operating characteristic curve (ROC) for each test were recorded and graphed. Our results for \texttt{MultiTok} were compared with BERT tokenization \cite{devlin2018bert}, GPT \cite{radford2019language} tokenization, and single-word tokenization where each unique word is considered as a token. For all these schemes, we use random vectors for embedding of each token.
We note that the BERT and GPT tokenizers are well-known off-the-shelve tokenizers utilized for LLM training today, and thus we targeted them as baselines in our experiments. Thus, for simulation results, if the proposed \texttt{MultiTok} tokenization strategy was able to perform better than these standards, it would be able to perform better than most other tokenization schemes. For this reason, an extensive comparison of more tokenization schemes, while interesting, would only provide marginal benefits.

Our first round of experiments are focused on studying the performance \texttt{MultiTok} algorithm for various sizes of the training window and the testing window. In the second experiment, we apply \texttt{MultiTok} encoding on the BERT and GPT tokens to have a cascade of two tokenization schemes, where \texttt{MultiTok} takes BERT or GPT tokens as units of data for compression. Finally, we apply the frequency-based post processing on the  \texttt{MultiTok} tokens for further improvements. 
We note that the title of rows in the table can be read as the type of tokenization, the percentage of training data to which the tokenizer is applied, and the training and testing window sizes.

\begin{figure*}[]
\centerline{\includegraphics[width=0.9\textwidth] {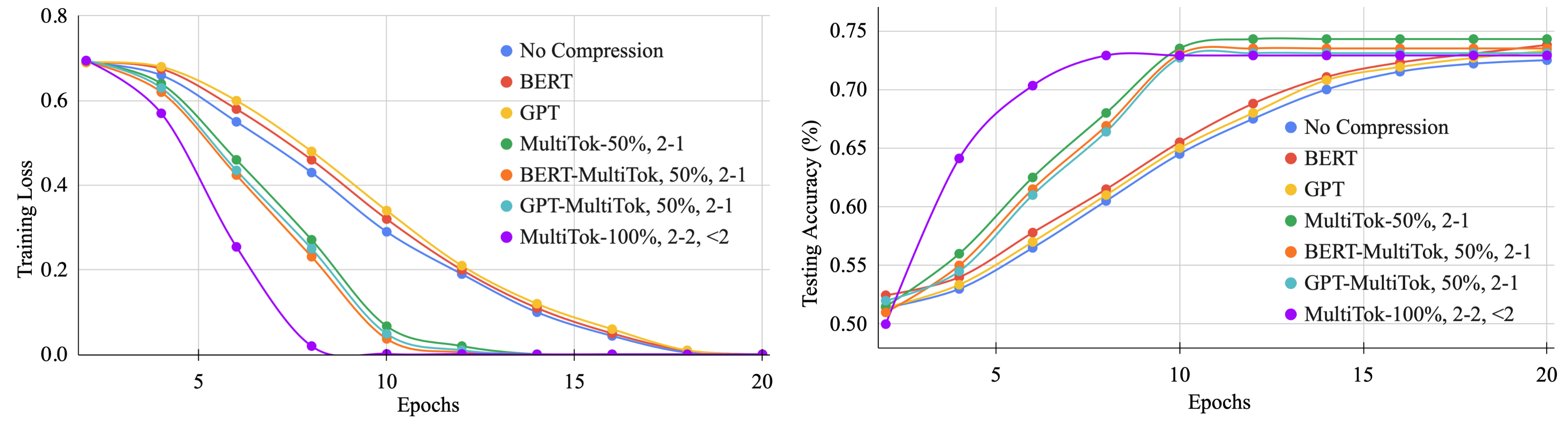}}
\caption{Training (left) \& Testing Graphs (right) for the IMDB dataset. }
\label{training}
\end{figure*} 

\begin{figure}
    \centering
    \includegraphics[width=0.95\linewidth]{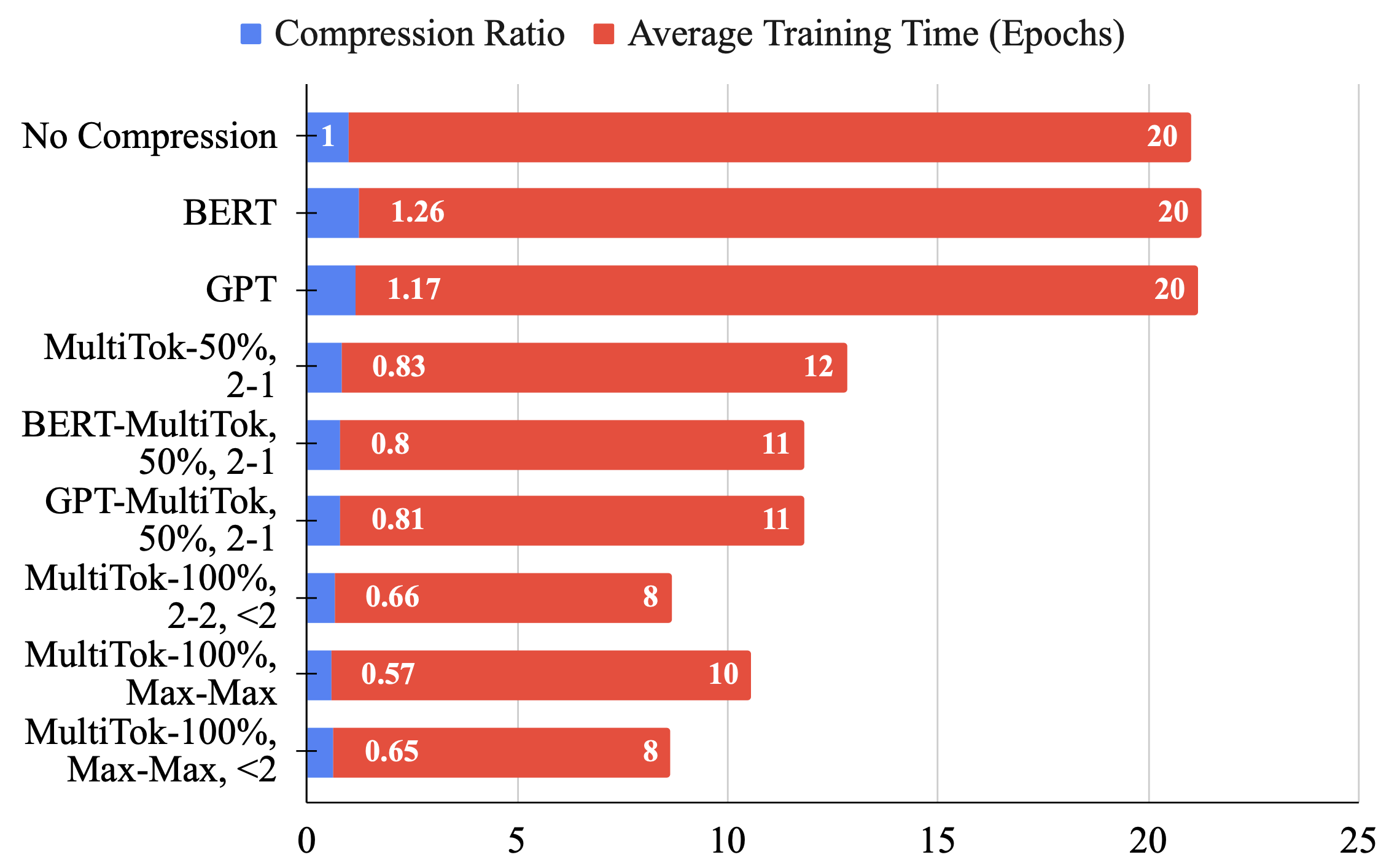}
    \caption{\texttt{MultiTok} demonstrates an average training time ($C(\epsilon) | \epsilon=0.01$) that is almost 2-3 times faster than the BERT \& GPT standards.}

    \label{fig:compression}
\end{figure}

\begin{table}
  \caption{Results on the IMDB dataset with 20+ trials}
  \label{tab:results}
\footnotesize
  \begin{tabular}{@{}c@{}ccc@{}}
    \toprule
    \textbf{Test}    & \textbf{Comp. ($r$)}    & \textbf{Accuracy} &\textbf{AUC}   \\
    \midrule
     BERT & 1.17 & 0.738 & 0.81 \\ 
     GPT & 1.26& 0.733 & 0.80 \\ 
     No Compression & 1 & 0.725 & 0.77 \\ \midrule
     {MultiTok}, 100\%, Max-Max & 0.57 & 0.648 & 0.69 \\ 
     {MultiTok}, 100\%, Max-2 & 0.57 & 0.643 & 0.69 \\ 
     {MultiTok}, 100\%, Max-1 & 0.57 & 0.50 & 0.5 \\
     {MultiTok}, 100\%, 2-2 & 0.6 & 0.701 & 0.75 \\ 
     {MultiTok}, 50\%, Max-Max & 0.8 & 0.661 & 0.72 \\ 
     {MultiTok}, 50\%, Max-1 & 0.8 & 0.718 & 0.77 \\ 
     {MultiTok}, 50\%, 2-2 & 0.83 & 0.706 & 0.76 \\ 
     {MultiTok}, 50\%, 2-1 & 0.83 & 0.743 & 0.82 \\ 
     {MultiTok}, 25\%, 2-1 & 0.93 & 0.728 & 0.78 \\ \midrule
     BERT-{MultiTok}, 50\%, 2-1 & 0.8 & 0.735 & 0.81\\ 
     BERT-{MultiTok}, 25\%, 2-1 & 0.91 & 0.753 & 0.83 \\
     GPT-{MultiTok}, 50\%, 2-1 & 0.81 & 0.731 & 0.79\\ 
     GPT-{MultiTok}, 33\%, 2-1 & 0.88 & 0.745 & 0.82 \\\midrule
     {MultiTok}, 100\%, 2-2, $\geq$2 & 0.66 & 0.729 & 0.79\\ 
     {MultiTok}, 100\%, Max-Max, $\geq$2 & 0.65 & 0.709 & 0.77 \\ \midrule
    \bottomrule
  \end{tabular}
\end{table}

\subsection{MultiTok Tokenization}
Table~\ref{tab:results} shows the results for tokenization schemes with different parameters. According to this table, the best performance is obtained by the presented \texttt{MultiTok} tokenization when the size of training and testing windows are $2$ and $1$, respectively, and the tokenization dictionary is applied to $50\%$ of samples in the training data. We remind that the testing window size $1$ means that during the inference from the trained downstream model, we used single-word tokens, and multi-word tokens were only used during training. In this scenario, \texttt{MultiTok} tokenization exhibits a slightly better performance compared to the standard BERT's performance ($0.743$ compared to $0.739$) on the IMDB dataset, while compressing the training data by $17\%$. 

As the amount of data that the \texttt{MultiTok} tokenization is applied to increases (the size of tokenized training data decreases), the performance of the downstream model decreases. Similarly, as the training window of the \texttt{MultiTok} tokenization scheme increases, the predictive performance of the downstream model decreases, as \texttt{MultiTok} groups tokens that rarely happen together in the training samples. Thus, there is an inverse relationship between the \texttt{MultiTok} tokenization amount and its training window; and the predictive performance of the downstream models. This is a reasonable observation as compression and performance are often competing goals.

\subsection{Combining MultiTok with Existing Tokenizers}
In this experiment, we used \texttt{MultiTok} tokenization to further compress the repetitive sequences within BERT and GPT tokens. This consequently reduces the training data size while achieving similar performance to that of normal uncompressed BERT \& GPT tokens. We observed that we could compress the training data that is tokenized via BERT by $26\%$ and via GPT by $38\%$ using \texttt{MultiTok}, while achieving the highest accuracies of any downstream model with $75.3\%$ and $74.5\%$ respectively. However, when we further reduced training data size, a higher compression is obtained at the cost of slight degradation in the performance. This suggests a notion of using a `golden ratio' of \texttt{MultiTok} compression to maximize efficiency and performance. This `golden ratio' may vary based on the repetition and succinctness found within different datasets. 

Next, we provide an empirical training convergence analysis, which shows that \texttt{MultiTok} not only performs as good as the BERT \& GPT standards in terms of accuracy and AUC, but also directly results in a faster training process for the downstream models. 

We observe in Fig.~\ref{training} that applying \texttt{MultiTok} tokenization to existing BERT and GPT tokens not only reduces the size of the training data, but also allows the model to train faster. Specifically, the combination of \texttt{MultiTok} and BERT or GPT tokenization achieves an average training time (as defined in Definition ~\ref{def:training-time}) of epoch 11 rather than epoch 20 (almost $2x$ faster) as shown in Fig. \ref{fig:compression}. Thus, these experiments suggest that \texttt{MultiTok} can be cascaded with existing tokenizers, like BERT and GPT, for smaller training data sizes, more efficient training, and improved downstream model performance.

\subsection{MultiTok Tokenization with Post-Processing}
Our results in this experiment indicates that the post processing of \texttt{MultiTok} tokenization with a frequency analysis component dramatically increases the performance of the resulting tokens. We remind that without this post processing, there is a significant decrease in the model's performance when the full capacity of \texttt{MultiTok} is utilized (when \texttt{MultiTok} is applied to the entire dataset with maximum window sizes), see Table \ref{tab:results}. However, by pruning less-utilized tokens, we were able to circumvent any impact to accuracy, as shown by the last set of experimental results. Specifically for experiment ``\texttt{MultiTok}, 100\%, 2-2, $\geq$2" (\texttt{MultiTok} is applied to the entire dataset with a training window size of 2), we see that by simply removing all multi-word tokens that only appear once in the training data, we get an accuracy of around $73\%$. While the training data size of the initial \texttt{MultiTok} experiment (\texttt{MultiTok}, 100\%, 2-2,) increases by $6 \%$, the resulting more succinct tokens provide an accuracy comparable to the BERT \& GPT baseline standards while still reducing the original training data size by $34\%$.

According to Fig.~\ref{training}, using these post-processed \texttt{MultiTok} tokens can result in a comparable accuracy up to $2.5x$ faster than when using BERT or GPT tokens. Most impressively, at epoch 8 in Fig.~\ref{training}, the loss of this stand-alone post-processed \texttt{MultiTok} experiment has almost achieved its stable minimum value, being at least 0.4 lower than the training loss for the baselines at the same epoch. In addition, as shown in Fig.~\ref{fig:compression}, the average training time for \texttt{MultiTok} based experiments is 8-12 epochs, which is significantly lower than the other counterpart tokenizers. Significantly, \texttt{MultiTok} has the lowest compression ratio out of the selected experiments, where the \texttt{MultiTok} tokens not only compressed the inputted training data by a factor of around $3$, but also helped the model learn the training data almost $2.5x$ faster. This confirms that with \texttt{MultiTok} tokenization, the model is able to learn the training data and existing tokens faster and more efficiently as opposed to the baseline tokenization schemes. 

\section{Conclusion \& Future Work}
We proposed \texttt{MultiTok}, a novel multi-word tokenizer which compresses repetitive words and phrases within a training dataset without harming the performance of downstream models. The proposed scheme can be used stand-alone and applied directly on words of text-based data, or it can be followed by an off-the-shelf tokenizer to improve efficiency and performance. Specifically, we demonstrate being able to compress training data by up to $33\%$ and train the downstream model $2.5x$ faster while maintaining comparable performance to the BERT and GPT standards. A promising next step is to extend \texttt{MultiTok} to be used for complex NLP tasks like text generation and even other data types like video. We hope that this paper marks the start of utilizing such data compression techniques to create more efficient LLMs of the future.

\bibliographystyle{IEEEtran}
\bibliography{references}
\end{document}